\pdfoutput=1

\documentclass[11pt]{article}

\usepackage[final]{acl}

\usepackage{times}
\usepackage{latexsym}

\usepackage[T1]{fontenc}

\usepackage[utf8]{inputenc}

\usepackage{microtype}

\usepackage{inconsolata}

\usepackage{graphicx}

\usepackage{algorithm}
\usepackage{algorithmic}
\usepackage{multirow}
\usepackage{pifont}
\usepackage{amsmath}
\usepackage{bm}
\usepackage{amssymb}
\usepackage{soul}
\usepackage{makecell}
\usepackage{subfigure}
\usepackage{enumitem}
\usepackage{xcolor}
\usepackage{colortbl}
\usepackage{booktabs}
\newcommand{\gyx}[1]{{\color{black}#1}}

\title{Graph-guided Cross-composition Feature Disentanglement for Compositional Zero-shot Learning}

\author{
 \textbf{Yuxia Geng\textsuperscript{1,2}},
 \textbf{Runkai Zhu\textsuperscript{2}},
 \textbf{Jiaoyan Chen\textsuperscript{3}},
 \textbf{Jintai Chen\textsuperscript{4}},
 Xiang Chen\textsuperscript{5}\thanks{Corresponding Author with \href{xiang_chen@nuaa.edu.cn}{xiang\_chen@nuaa.edu.cn}},
 \\
 \textbf{Zhuo Chen\textsuperscript{6}},
 \textbf{Shuofei Qiao\textsuperscript{6}},
 \textbf{Yuxiang Wang\textsuperscript{2}},
 \textbf{Xiaoliang Xu\textsuperscript{2}},
 \textbf{Sheng-Jun Huang\textsuperscript{5}}
\\
\normalsize{
 \textsuperscript{1}PowerChina Huadong Engineering Corporation Limited
 \textsuperscript{2}Hangzhou Dianzi University
 }
  \\
  \normalsize{
 \textsuperscript{3}The University of Manchester
 \textsuperscript{4}The Hong Kong University of Science and Technology (Guangzhou)
 }
 \\
 \normalsize{
 \textsuperscript{5}Nanjing University of Aeronautics and Astronautics
 \textsuperscript{6}Zhejiang University
 }
}

\begin{document}
\maketitle
\begin{abstract}
Disentanglement of visual features of primitives (i.e., attributes and objects) has shown exceptional results in Compositional Zero-shot Learning (CZSL). 
However, due to the feature divergence of an attribute (resp. object) when combined with different objects (resp. attributes), it is challenging to learn disentangled primitive features that are general across different compositions.
To this end, we propose the solution of \textit{cross-composition feature disentanglement}, which takes multiple primitive-sharing compositions as inputs and constrains the disentangled primitive features to be general across these compositions.
More specifically, we leverage a compositional graph to define the overall primitive-sharing relationships between compositions, and build a task-specific architecture upon the recently successful large pre-trained vision-language model (VLM) CLIP, with dual cross-composition disentangling adapters (called L-Adapter and V-Adapter) inserted into CLIP's frozen text and image encoders, respectively.
Evaluation on three popular CZSL benchmarks shows that our proposed solution significantly improves the performance of CZSL, and its components have been verified by solid ablation studies.
Our code and data are available at: \url{https://github.com/zhurunkai/DCDA}.
\end{abstract}

\section{Introduction}
Compositional Zero-shot Learning (CZSL) aims to recognize novel attribute-object compositions by disentangling visual primitives from seen combinations, a capability crucial for scaling visual recognition systems \cite{misra2017red}. For instance, a model trained on \textit{red tomato} and \textit{green apple} should infer \textit{green tomato} through primitive recombination, despite never encountering this specific composition. This paradigm not only enables zero-shot generalization to exponentially many combinations \cite{chen2023zero}, but also advances vision-language understanding by requiring precise feature decomposition aligned with textual semantics \cite{chen2025knowledge}.

\begin{figure}[t]
  \centering  \includegraphics[width=\linewidth]{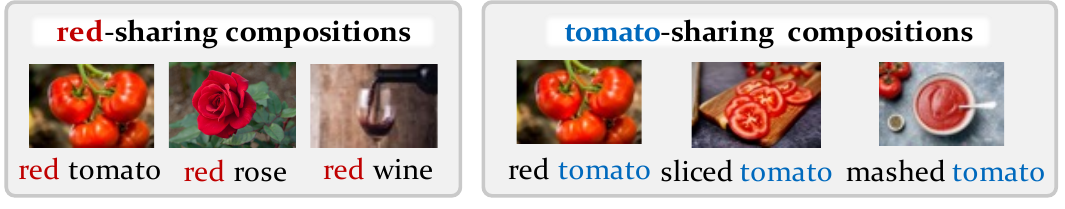}
\caption{Examples of divergent visual features of a primitive (e.g., an attribute \textit{red} or an object \textit{tomato}) across different compositions.
}
\label{fig:composition_examples}
\end{figure}

Early CZSL approaches establish shared embedding spaces to compare image features with composition embeddings \cite{wei2019adversarial,naeem2021learning,mancini2022learning}. Recent advances leverage CLIP's visual-semantic alignment from large-scale pretraining \cite{radford2021learning,csp2023}, yet face inherent challenges: Attribute-object primitives exhibit strong visual entanglement—consider how \textit{red} permeates all pixels of a \textit{red tomato}. This entanglement hinders both primitive alignment and novel composition generalization. Current CLIP-based solutions address this through either disentangled text prompts \cite{dfsp2023,hpl2023} or vision adapters \cite{caila2024,huang2024troika,li2024context}, but crucially overlook the \textbf{diversity} of primitive manifestations across compositions. As Figure~\ref{fig:composition_examples} illustrates, the visual realization of \textit{red} varies significantly when combined with different objects (e.g., \textit{tomato} vs. \textit{wine}), exhibiting divergent color tones and spatial distributions. 

\begin{figure*}[t]
\centering
\subfigure{
\begin{minipage}[c]{0.5\linewidth}
\centering
\includegraphics[width=5.5cm]{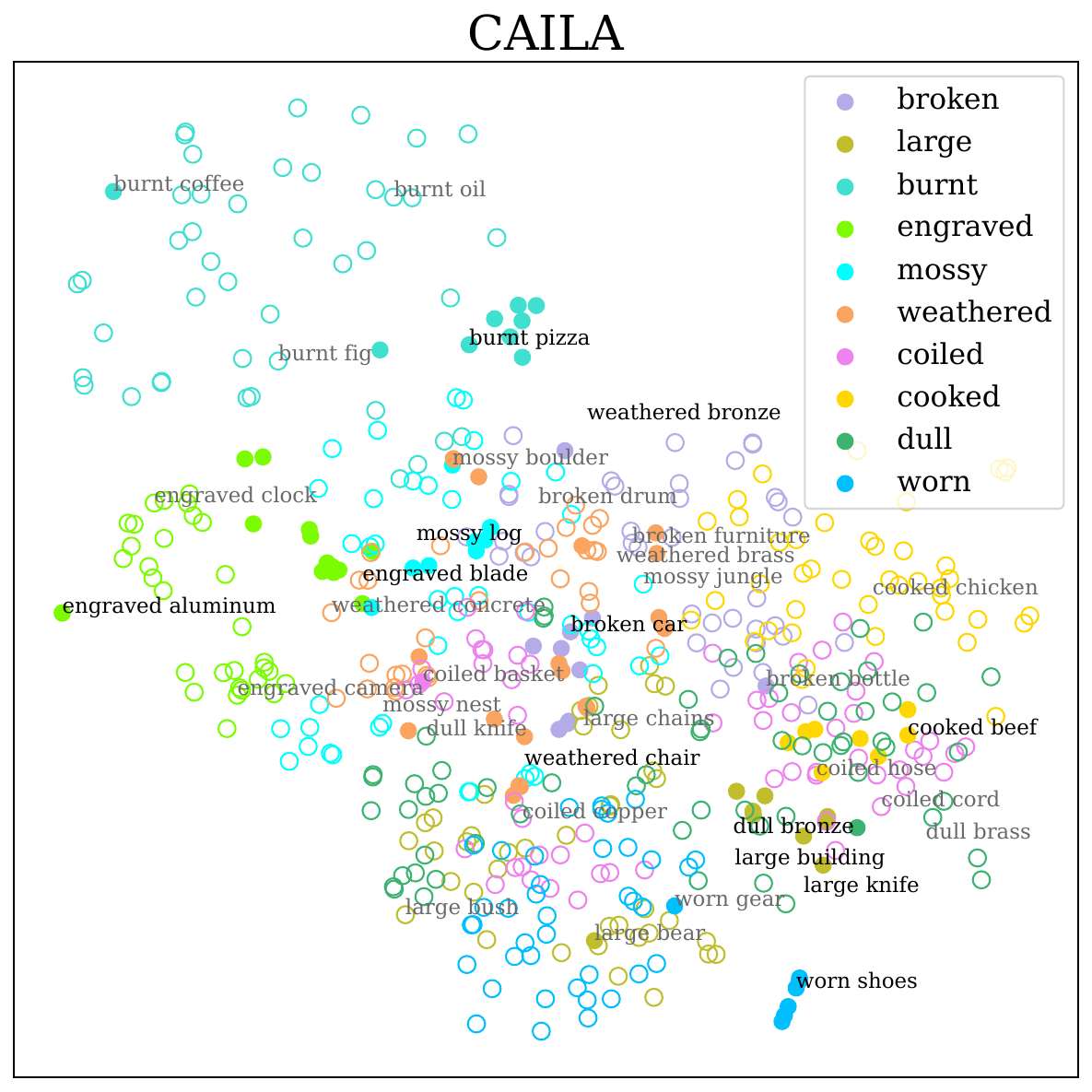}
\end{minipage}%
}%
\subfigure{
\begin{minipage}[c]{0.5\linewidth}
\centering
\includegraphics[width=5.5cm]{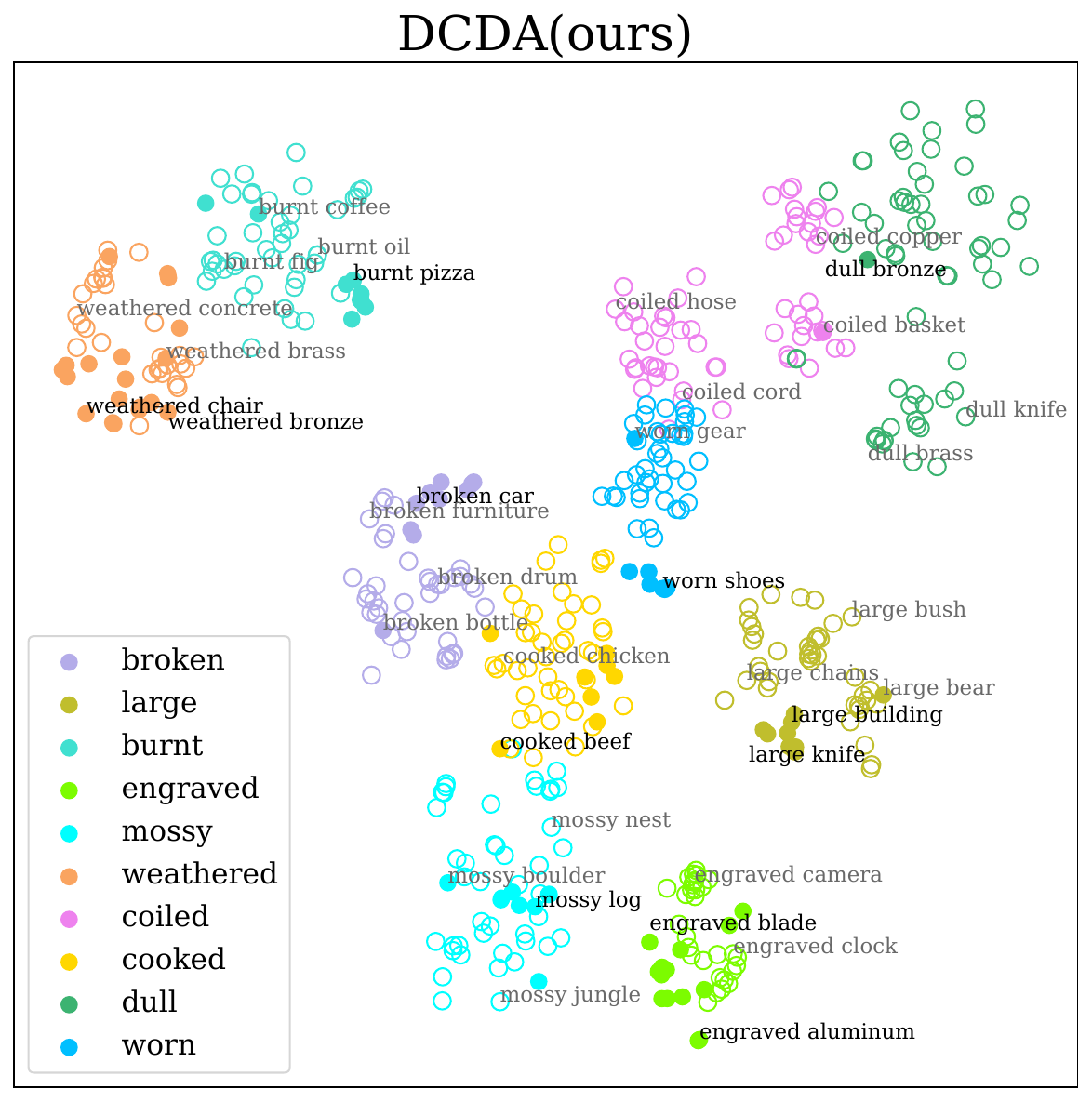}
\end{minipage}%
}%
\centering
\caption{$t$-SNE visualizations of disentangled attribute representations of images in the test set of MIT-States, learned by CAILA \cite{caila2024} and our DCDA. Solid and hollow circles represent images of seen and unseen compositions, respectively. Best viewed in color.
}\label{fig:disentangled_attribute_visualization}
\end{figure*}

To quantify this challenge, we conduct feature-space analysis on MIT-States \cite{isola2015discovering} using CAILA \cite{caila2024}, the current SOTA method. As visualized in Figure~\ref{fig:disentangled_attribute_visualization} (left), disentangled attributes like \textit{broken} (purple circles) exhibit two critical limitations: (1) Features from the same attribute scatter widely with overlapping clusters (e.g., \textit{broken} intermixed with \textit{cooked} and \textit{mossy}), indicating poor intra-class consistency; (2) This dispersion directly impacts generalization—compositions containing unseen \textit{broken} objects become indistinguishable due to the attribute's non-discriminative embeddings.

\begin{figure*}[]
  \centering  \includegraphics[width=1.0\linewidth]{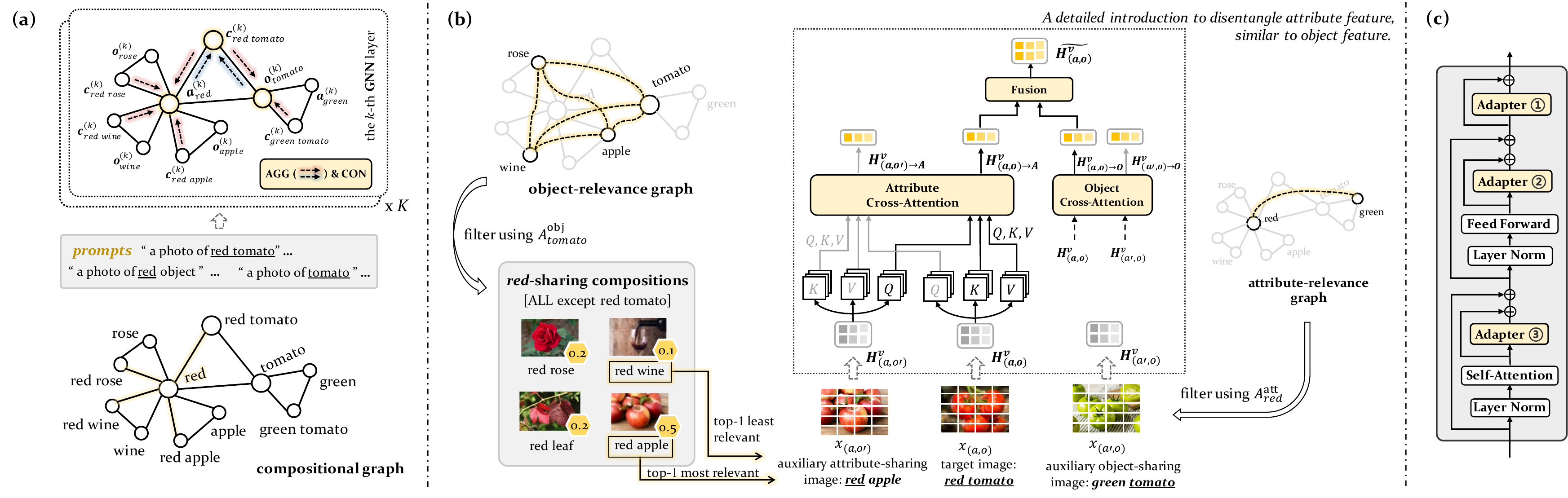}
\caption{Overview of DCDA during training: (a) The L-Adapter built upon the composition graph and GNN module; (b) The V-adapter built upon the cross-attention and attribute/object relevance-guided sampling strategy; (c) An illustration of the candidate position for inserting adapters in a transformer block. We take the learning of \textit{red tomato} as an example.}
\label{fig:framework}
\end{figure*}

To overcome these limitations, we propose cross-composition feature aggregation through graph-guided learning. Our key insight is that
effective primitive disentanglement requires cross-composition feature aggregation. Drawing inspiration from compositional graphs \cite{naeem2021learning}, we construct a tripartite graph (Figure~\ref{fig:framework}a) connecting attributes, objects, and their compositions. This graph enables our \textbf{D}ual \textbf{C}ross-composition Feature \textbf{D}ecomposing \textbf{A}dapters (\textbf{DCDA}), which enhance CLIP's text and image encoders through complementary strategies:
\textbf{(1) L-Adapter} propagates textual features across compositionally related nodes via GNNs, consolidating attribute/object semantics from multiple contexts.
\textbf{(2) V-Adapter} employs cross-attention between \gyx{primitive-sharing} images (e.g., \textit{red tomato} and \textit{red wine}) to extract invariant visual patterns, augmented by a novel \gyx{sampling strategy that weightedly selects compositions according to the attribute/object co-occurrence degrees (derived from our graph).}
When integrating L-Adapters and V-Adapters into multiple layers of CLIP's text and image encoders, we retain the original parameters of CLIP to avoid overfitting, but inject the task-specific knowledge.
Our contributions can be summarized below:
\begin{itemize}[leftmargin=0.6cm]
    \item DCDA is the first systematic approach for cross-composition feature disentanglement in CLIP-based CZSL, explicitly addressing primitive diversity.
    \item Dual adapter architecture leveraging compositional graphs for text-side aggregation and vision-side contrastive attention, enabling discriminative yet generalizable primitive representations.
    \item DCDA achieves great performance on MIT-States and UT-Zappos (closed/open world), with 5.1\%/7.3\% gains over CAILA \gyx{method}\footnote{Visualizations confirm tighter clustering of disentangled attributes (Figure~\ref{fig:disentangled_attribute_visualization}, right), with extended object analyses in Appendix~\ref{sec:appendix_case_study}.}.

\end{itemize}

\section{Related Work}

\textbf{Conventional CZSL} methods are roughly divided into two groups. One is classifier-based which first trains two separate classifiers to predict an input image's attribute and object labels, respectively, and then combines them to predict the compositional labels \cite{misra2017red,nagarajan2018attributes}.
The subsequent works further enhance the dependence of the attribute and object in a composition \cite{li2020symmetry,li2022siamese,wang2023learning}.
The other group is embedding-based which jointly represents attributes and objects to capture the dependence, and then aligns them with the images in a shared embedding space \cite{wei2019adversarial,karthik2022kg,geng2021ontozsl}.
In particular, \cite{naeem2021learning} learn the joint representation through graph convolutional networks.
There are also some works concerning the disentanglement of attribute and object features in the visual space \cite{saini2022disentangling,hao2023learning,kim2023hierarchical,mkgformer,unihd} or the label space \cite{geng2022disentangled,knowprompt,dqset}.
However, all these methods have to learn the alignment between image features and text embeddings from scratch and are prone to overfit to the seen compositions.
It is expected to derive pre-trained alignment knowledge from VLMs.

\noindent\textbf{CLIP-based CZSL.} After pre-training using 400 million image-text pairs, CLIP can be applied to any visual classification task without fine-tuning by setting prompts like ``a photo of [class]'', where ``[class]'' is filled with the name of the class to be recognized.
\cite{csp2023} then had the first attempt to design prompt ``a photo of [attribute] [object]'' for CZSL, where ``[attribute] [object]'' are tunable tokens to teach CLIP how to compose attributes and objects.

To stress the roles of individual primitives, \cite{hpl2023} additionally set an attribute and an object prompt with only ``[attribute]'' or ``[object]'' tunable; \cite{dfsp2023} make the whole prompt trainable and fuses the decomposed text features with the encoded (entangled) image features through a cross-modal fusion module.
Different from these works focusing on optimizing the prompts, \cite{caila2024} propose to insert trainable adapters inside the frozen transformer layers to decompose and recompose the attribute and object features.
With disentangled primitive features, \cite{huang2024troika} establish three prediction branches and pulls a static class prompt to its dynamic images via a cross-modal traction module. \cite{li2024context} investigate the relative specificity of attributes when paired with different objects.
In contrast to these methods, our method DCDA is more generalizable, with cross-composition knowledge injected and dual adapters inserted in CLIP's image and text encoders.


\section{Methodology}

\textbf{CZSL Task Formulation.}
Let $\mathcal{D}_{tr}=\{(x,c)|x\in\mathcal{X}_s, c \in \mathcal{C}_s\}$ be the training set, where $\mathcal{X}_s$ contains the training images and $\mathcal{C}_s$ is a set of seen compositional labels that are available during training. 
Each label is a tuple $c=(a,o)$ of an attribute class $a \in \mathcal{A}$ and an object class $o \in \mathcal{O}$.
After training, the CZSL model can predict images of a set of new compositions $\mathcal{C}_u$ that are unseen during training, with $\mathcal{C}_u \cap \mathcal{C}_s = \emptyset$. 
Following previous works, we study generalized CZSL \cite{purushwalkam2019task}, where images of seen and unseen compositions are both tested and the candidate label space includes both seen and unseen labels. The test set is thus denoted as $\mathcal{D}_{te}=\{(x,c)|x\in \mathcal{X}_{te}, c \in \mathcal{C}_{te}\}$, where $\mathcal{X}_{te} = \mathcal{X}_u \cup \mathcal{X}'_s$ with $\mathcal{X}'_s \cap \mathcal{X}_s = \emptyset$, and $\mathcal{C}_{te} = \mathcal{C}_u \cup \mathcal{C}_s$.
Notably, $\mathcal{C}_s$ and $\mathcal{C}_u$ share the same attribute set $\mathcal{A}$ and object set $\mathcal{O}$, CZSL assumes that each $a$ and $o$ has been trained before testing and only the composition $(a,o) \in \mathcal{C}_u$ is novel.

\noindent\textbf{Overview.}
In the following, we will first introduce the details of \textbf{L-Adapters} for the language side and \textbf{V-Adapters} for the vision side, and then introduce how to integrate them into the frozen CLIP encoders for CZSL.  
As shown in Figure \ref{fig:framework}(c), the adapters are inserted into CLIP's intermediate computational units such as self-attention layers or feed-forward layers.
This means the input of each adapter is the output of CLIP's one computational unit, and its output is the input of CLIP's next computational unit.
We use $\bm{H}^t \in \mathbb{R}^{l \times d}$ and $\bm{H}^v \in \mathbb{R}^{l' \times d'}$ to denote the output of CLIP's one specific computational unit in the text and image encoders, respectively, where $l$ (resp. $l'$) is the length of a tokenized input text (resp. image), $d$ and $d'$ is the hidden state size of each token.

\subsection{The design of L-Adapter}
Each L-Adapter is built upon a compositional graph for representing the global compositional relationships among attributes, objects and compositions, and a GNN module for propagating and aggregating features among them to realize the cross-composition learning of textual primitive features.

We first define the compositional graph. It consists of $N=|\mathcal{A}| + |\mathcal{O}| + |\mathcal{C}'|$ nodes, including all the attributes, all the objects, and the compositions in the current computation (i.e., $\mathcal{C}' =\mathcal{C}_s$ for training and $\mathcal{C}' =\mathcal{C}_{te}$ for testing).
Given these nodes, for each $c=(a,o)$, we connect $(a,o)$, $(a, c)$ and $(o,c)$ to form a triangle in the graph, as shown in Figure~\ref{fig:framework}(a).
For simplicity and efficiency, we keep all graph edges unweighted and undirected as in \cite{naeem2021learning} and obtain a symmetric adjacency matrix $A\in \mathbb{R}^{N \times N}$ to store the graph structure, $A_{ij} = 1$ if there is a connection between node $i$ and $j$ otherwise $A_{ij}=0$. 

To obtain the initial representations of graph nodes, we first define individual prompts for attributes and objects as ``a photo of [attribute] object'' and ``a photo of [object]'' besides the composition prompts ``a photo of [attribute] [object]'', then for each $c=(a,o)$, we feed three prompts into CLIP's text encoder to output three hidden states $\bm{H}^t_a$, $\bm{H}^t_o$ and $\bm{H}^t_c$ for $a,o,c$, respectively, and finally extract the embeddings of the special token \texttt{[EOT]} in the prompts as the initial features $\{\bm{h}_i^{t}\}_{i=1}^{N}$ of graph nodes, with $\bm{h}^t_i = \bm{H}_{i, \texttt{[EOT]}}^{t}$ and $\bm{h}_i^t \in \mathbb{R}^d$. In this way, the text feature of each primitive is naturally disentangled from the composition one.

Next, we exploit multiple GNN layers to propagate features among graph nodes following the graph structure defined in $A$. 
Formally, as Figure~\ref{fig:framework}(a)'s top shows, for each $c=(a,o)$, three identical \textbf{AGG} functions are parallelly applied to aggregate neighborhood features for nodes $a,o,c$ at the $k$-th GNN layer, $k \in \{0,..., K-1\}$, as:
\begin{align}
    & \bm{a}^{(k)}_{\mathcal{N}_a} = \text{AGG}^{(k)} (\{\bm{c}^{(k)}_{i}| c_i \in \mathcal{N}_a^c\}, \{\bm{o}^{(k)}_{i}| o_i \in \mathcal{N}_a^o\})
\\
&  \bm{o}^{(k)}_{\mathcal{N}_o} = \text{AGG}^{(k)} (\{\bm{c}^{(k)}_{j}| c_j \in \mathcal{N}_o^c\}, \{\bm{a}^{(k)}_{j}| a_j \in \mathcal{N}_o^a\})
 \\
  &  \bm{c}^{(k)}_{c=(a,o)} = \text{AGG}^{(k)} (\bm{a}^{(k)}, \bm{o}^{(k)})
\end{align}
where $\mathcal{N}_a^c$ (resp. $\mathcal{N}_o^c$) denotes the composition neighbor set of $a$ (resp. $o$) on the graph, and $\mathcal{N}_a^o$ (resp. $\mathcal{N}_o^a$) includes the objects (resp. attributes) that compose the compositions in $\mathcal{N}_a^c$ (resp. $\mathcal{N}_o^c$) together with $a$ (resp. $o$). 
Considering the example in Figure~\ref{fig:framework}(a) where $c=(a,o)$ is \textit{red tomato}, $\mathcal{N}_a^c$ includes compositions like \textit{red apple} and $\mathcal{N}_a^o$ includes objects like \textit{apple}.
While the neighbor set of each $c$ only contains its primitives $a$ and $o$. $\bm{a}^{(k)}$ is the input feature of $a$ at the $k$-th layer, and is updated using \textbf{CON} function to obtain the $k$-th-layer output as $\bm{a}^{(k+1)} = \text{CON}(\bm{a}^{(k)}_{\mathcal{N}_a}, \bm{a}^{(k)})$, similar for $o$ and $c$ with outputs $\bm{o}^{(k+1)}$ and $\bm{c}^{(k+1)}$.
The input feature of $a,o,c$ at the first layer of GNN is the initialized node feature, e.g., $\bm{c}^{(0)} = \bm{h}_c^{t}$.
The output features of $a,o$ after $K$ GNN layers $\bm{a}^{(K)}, \bm{o}^{(K)}$, which have already fused their neighboring compositions' features, and $c$'s output feature $\bm{c}^{(K)}$, which has aggregated the updated features of $a$ and $o$, are the final output of one L-Adapter,
and will be inputted into the next computation unit of CLIP for the latter computation.

\subsection{The design of V-Adapter}

Since attributes and objects are highly entangled within the input image, we cannot build the same computational graph in V-Adapters as that in L-Adapters.
Targeting this, we leverage the cross-attention over primitive-sharing image pairs to extract cross-composition-sharing primitive features, and design a primitive relevance-guided sampling strategy to introduce more valid primitive-sharing compositions. We take disentangling attribute features from an input image as an example, object features are processed similarly. 

Specifically, as shown in Figure~\ref{fig:framework}(b), given a target image $x_{(a,o)}$ to predict, which is labeled by $c=(a,o)$, we first randomly sample an auxiliary composition that shares the same attribute as $x_{(a,o)}$ but has different object $o'$, and select one of its images $x_{(a,o')}$ as an auxiliary image\footnote{Regarding the absence of attribute-sharing compositions or images, let’s consider an input image of \textit{red apple}, if there are no other red objects in the dataset, we take \textit{red apple} itself as the auxiliary composition and sample one of its samples (excluding the input one) as the auxiliary image. And, if there are no other \textit{red apple} images, we treat the input image itself as the auxiliary image.}.
Then, we feed these two images into CLIP to output two hidden states $\bm{H}_{(a,o)}^v$ and $\bm{H}_{(a,o')}^v$ and compute the cross-attention as, with $\bm{H}_{(a,o')}^v$ as the query and $\bm{H}_{(a,o)}^v$ as the key and value:
\begin{gather}
    \text{CrossAttention} (\bm{Q,K,V}) = softmax (\tfrac{\bm{Q}\bm{K}^T}{\sqrt{d'}}) \bm{V}
    \\
    \bm{Q} = \bm{H}_{(a,o')}^v\bm{W}_Q,
    \bm{K} = \bm{H}_{(a,o)}^v\bm{W}_K,
    \bm{V} = \bm{H}_{(a,o)}^v\bm{W}_V
\end{gather}
where $\bm{W}_{\{Q,K,V\}} \in \mathbb{R}^{d' \times d'}$ are three linear transformation matrices for flexible computation.
We can see that every output embedding is a weighted sum of the value embeddings, and the weights are calculated by the similarity of the query and the key.
By setting query as $\bm{H}_{(a,o')}^v$, we can refine $\bm{H}_{(a,o)}^v$ to keep features that are more specific to $a$, as well as attribute features that are general across these two compositions.
We also swap the query and the value (also key) to refine $\bm{H}_{(a,o')}^v$.
The output of cross-attention is thus denoted as $\bm{H}_{(a,o)\rightarrow A}^{v}$ and $\bm{H}_{(a,o')\rightarrow A}^v$.
A feed-forward layer is also added after the cross-attention.


Notably, the above cross-attention can only process two attribute-sharing compositions at one time. To introduce more compositions to learn more general attribute features, the model relies on the batched data and random sampling to switch the auxiliary composition. However, when an attribute is diverse with extensive composition neighbors, e.g., \textit{red} or \textit{broken}, which also means extensive candidate auxiliary compositions, the model requires more switches to traverse them, leading to inferior overall performance as shown in Table~\ref{tab:overall_results}.
To balance the switch times of attributes with different numbers of neighbors, we propose to select some representative compositions instead of all the compositions as the candidates, and for a target image $x_{(a,o)}$, the top-$n$ objects that are most and least relevant to the target object $o$ are paired with the target attribute $a$ to serve as the representative auxiliary compositions, while the relevance between two objects can be determined by the number of common attributes that co-occur in their associated compositions in the training set; for example, if there are compositions \textit{red tomato} and \textit{red apple} in the training set, \textit{red} is a common attribute of the objects \textit{tomato} and \textit{apple}.

To this end, we refer to the attribute-object edges in the training compositional graph to first create an \textbf{att-obj} graph with structure matrix $A^{\text{att-obj}} \in \mathbb{R}^{|\mathcal{A}| \times |\mathcal{O}|}$, where $A^{\text{att-obj}}_{i,j}=1$ means attribute $i$ and object $j$ form a valid seen composition while $0$ not. Then, an object relevance graph can be created and its structure matrix is found as $A^{\text{obj}} = (A^{\text{att-obj}})^T A^{\text{att-obj}}$ with size $|\mathcal{O}| \times |\mathcal{O}|$, where $A^{\text{obj}}_{i,j}$ represents the number of common attributes between $i$-th and $j$-th objects, large numbers mean higher relevance. With this relevance graph, for target image $x_{(a,o)}$, and all of its $a$-sharing compositions besides $(a,o)$, denoted as $\{(a, o')\}$, we next refer to $A^{\text{obj}}_o$ to obtain the relevance scores of objects in $\{o'\}$ w.r.t $o$, and select the compositions whose objects have top-$n$ maximum and top-$n$ minimum non-zero scores as the representative compositions. Figure \ref{fig:framework}(b) presents a running example of this procedure.
Finally, we perform a weighted random sampling over these representative compositions, i.e., the probability for selecting $(a,o')$ is determined by the normalized relevance score between $o'$ and $o$, more details are attached in Appendix~\ref{sec:appendix_method_sampling_prob}.

The same applies to the output of the object cross-attention $\bm{H}_{(a,o)\rightarrow O}^{v}$, which is learned from a set of representative $o$-sharing compositions selected by referring to the attribute relevance matrix $A^{\text{att}} = A^{\text{att-obj}} (A^{\text{att-obj}})^T$ and its row value $A^{\text{att}}_a$.
In this way, we learn the cross-composition-sharing primitive features of an input image, and have the updated image features: $\bm{\widetilde{H}}_{(a,o)}^v = \bm{H}_{(a,o)\rightarrow A}^v + \bm{H}_{(a,o)\rightarrow O}^v$, which will be the final output of one V-Adapter together with $\bm{H}_{(a,o)\rightarrow A}^v$ and $\bm{H}_{(a,o)\rightarrow O}^v$.

\subsection{Integrating Adapters into CLIP}\label{sec:insert_adapters}
Given a single L-Adapter and V-Adapter, we next present how to insert them into two connected computation units in CLIP.
Inspired by ViT \cite{dosovitskiy2020image} which tends to learn general features at lower layers and learn specific features at higher layers, we add our adapters starting from the top transformer blocks.
After preliminary validations on MIT-States (see our Ablation Study results for more), 
we decided to \textit{i)} insert adapters at the last three transformer blocks of both text and image encoders, and \textit{ii)} add L-Adapters behind the self-attention layer and feed-forward layer in each language transformer block (i.e., the positions \ding{174} and \ding{173} in Figure~\ref{fig:framework}(c)), and attach V-Adapters after the whole vision transformer block (i.e., the position \ding{172} in Figure~\ref{fig:framework}(c)).
Moreover, we build a skip connection between the input and the output of each adapter before inputting it into the next computation unit of CLIP.
We use CLIP's pre-trained word embeddings to initialize each word in our prompts, including the prefix words ``a photo of'', and keep these word embeddings trainable for capturing more task-specific knowledge.

\subsection{Training and Inference}
At the last transformer blocks, we obtain the output of L-Adapter and V-Adapter after skip connection, denoted as $\bm{\widehat{H}}^v$ and $\bm{\widehat{H}}^t$, based on this, we extract the embeddings of \texttt{[EOT]} token, i.e., $\bm{\widehat{h}}^{v} (=\bm{\widehat{H}}^{v}_{\texttt{[EOT]}})$ and $\bm{\widehat{h}}^{t} (=\bm{\widehat{H}}^{t}_{\texttt{[EOT]}})$, to measure the compatibility of visual and textual features.
Formally, for an input image $x_i$ and a training composition $c_i=(a_i,o_i)$, we compute the compatibility score as:
\begin{equation}\label{eq:score}
\begin{aligned}
s(x_i,c_i=(a_i,o_i)) &= \alpha \ [ \bm{\widehat{h}}^{v}_{i} \cdot \bm{\widehat{h}}^{t}_{c_i}] \\ &+ \beta \  [\bm{\widehat{h}}^{v}_{i \rightarrow A} \cdot \bm{\widehat{h}}^{t}_{a_i}] \\ &+ \gamma \ [\bm{\widehat{h}}^{v}_{i \rightarrow O} \cdot \bm{\widehat{h}}^{t}_{o_i}]
\end{aligned}
\end{equation}
where we also measure the individual primitive compatibility via the second and the third items as if one image belongs to an attribute-object pair, its disentangled attribute and object features also belong to the corresponding primitive labels.
We use three learnable parameters $\alpha, \beta, \gamma$ to balance the overall score.
$\cdot$ denotes the dot-product similarity.

We optimize these adapters and trainable token embeddings by minimizing the cross-entropy loss on the training set $\mathcal{D}_{tr}$ with seen compositions from $\mathcal{C}_s$, with $\tau$ as the temperature widely used in CLIP, and $c_i=(a_i,o_i)$ as the ground-truth label: 
\begin{equation}
    \mathcal{L} = - \frac{1}{|\mathcal{D}_{tr}|} \sum_{x_i \\ \in \mathcal{D}_{tr}} log \frac{e^{[ s(x_i,c_i=(a_i,o_i)) / \tau ]}}{\sum_{c_j \in \mathcal{C}_s} e^{[s(x_i,c_j=(a_j,o_j)/ \tau)]}}
\end{equation}

During inference, for each testing image $x_t$, we estimate the compatibility score between $x_t$ and each testing composition $c=(a,o)$ from $\mathcal{C}_{te}$ as Equation~\ref{eq:score}. The composition that has the highest compatibility score is the predicted label.
Besides, since we have no idea about the attribute and object labeled for $x_t$, we take itself as the primitive-sharing images to compute $\bm{H}_{t\rightarrow A}^v $ and $\bm{H}_{t\rightarrow O}^v$.

\begin{table*}[htbp]
\centering
\resizebox{0.95\linewidth}{!}{
\begin{tabular}{c|l|cccc|cccc|cccc}
\hline
\multicolumn{1}{c|}{\multirow{2}{*}{\textbf{Setting}}}
& 
\multicolumn{1}{c|}{\multirow{2}{*}{\textbf{Methods}}}
&
\multicolumn{4}{c|}{\textbf{MIT-States}} & \multicolumn{4}{c}{\textbf{UT-Zappos}} 
& \multicolumn{4}{c}{\textbf{C-GQA}}
\\
& &  S &  U & H & AUC &   S &  U & H & AUC &  S &  U & H & AUC
\\
\hline
\multicolumn{1}{c|}{\multirow{10}{*}{Closed World}}
& CGE 
& 32.8 & 28.0 & 21.4 & 6.5
& 64.5 & 71.5 & \textbf{60.5} & 33.5
& 33.5 & 15.5 & 16.0 & 4.2
\\
& ADE
& -- & -- & -- & --
& -- & -- & -- & --
& 35.0 & 17.7 & 18.0 & 5.2 
\\
& 
CLIP 
& 30.2 & 46.0 &26.1 & 11.0
& 15.8 & 49.1 & 15.6 & 5.0
& 7.5 & 25.0 & 8.6 & 1.4
\\
& 
CSP 
& 46.6 & 49.9 & 36.3 & 19.4
& 64.2 & 66.2 & 46.6 & 33.0
& 28.8 & 26.8 & 20.5 & 6.2
\\
& 
HPL 
& 47.5& 50.6 & 37.3 & 20.2
& 63.0 & 68.8 & 48.2 & 35.0
& 30.8 & 28.4 & 22.4 & 7.2
\\
& 
DFSP
& 46.9 &  52.0 & 37.3 & 20.6
& 66.7 & 71.7 & 47.2 & 36.0
& \textit{37.3} & \textit{26.1} & \textit{23.5} & \textit{8.2}
\\
& 
CAILA 
& 51.0 & 53.9 & 39.9 & 23.4
& 67.8 & \underline{74.0} & 57.0 & \underline{44.1}
& \textbf{\textit {40.4}} & \underline{\textit {28.6}} & \textbf{\textit {26.1}} & \textbf{\textit {9.9}}
\\
& 
Troika 
&  49.0 & 53.0 & 39.3 & 22.1 
&  66.8 & 73.8 & 54.6 & 41.7 
&  \textit{38.0} & \textit{28.4} & \textit{\underline{25.3}} & \textit{9.2} 
\\
\rowcolor{gray!15}
\cellcolor{white}{} & DCDA[RD]
& 42.2 & 46.7 & 32.8 & 16.2
& 64.7 & 71.5 & 54.5 & 40.1
& \textit{\underline{39.8}} & \textit{25.3} & \textit{23.9} & \textit{8.5}
\\
\rowcolor{gray!15}
\cellcolor{white}{} & DCDA[PRG]
& \textbf{57.3} & \underline{55.1} & \textbf{43.2} & \underline{26.9}
& \underline{68.7} & 72.4 & 56.5 & 43.0
& \textit{39.1} & \textit{26.7} & \textit{24.5} & \textit{8.9}
\\
\rowcolor{gray!15}
\cellcolor{white}{} & DCDA[PRG+N]
& \underline{57.1}& \textbf{55.5}& \underline{43.1}&\textbf{27.0}
& \textbf{69.1}& \textbf{74.1}& \underline{57.2}&\textbf{44.2}
& \textit{38.5} & \textit{\textbf{28.8}} & \textit{\underline{25.3}} & \textit{\underline{9.4}}
\\
\hline
\multicolumn{1}{c|}{\multirow{10}{*}{Open World}}
& CGE 
&32.4 & 5.1 & 6.0 & 1.0
& 61.7 & 47.7 & 39.0 & 23.1
& 32.1 & 1.8 & 2.9 & 0.47
\\
& ADE
& -- & -- & -- & --
& -- & -- & -- & --
& 35.1 & 4.8 & 7.6 & 1.42
\\
& 
CLIP 
& 30.1 & 14.3 & 12.8 & 3.0
& 15.7 & 20.6 & 11.2 & 2.2
& 7.5 & 4.6 & 4.0 & 0.27
\\
& 
CSP 
& 46.3 & 15.7 & 17.4 & 5.7
& 64.1 & 44.1 & 38.9 & 22.7
& 28.7 & 5.2 & 6.9 & 1.20
\\
& 
HPL 
& 46.4 & 18.9 & 19.8 & 6.9
& 63.4 & 48.1 & 40.2 & 24.6 
& 30.1 & 5.8 & 7.5 & 1.37
\\
& 
DFSP
& 47.5 & 18.5 & 19.3 & 6.8
& 66.8 & 60.0 & 44.0 & 30.3
& \textit{35.0} & \textit{4.9} & \textit{7.3} & \textit{1.42}
\\
& 
CAILA 
& 51.0 & 20.2 & 21.6 & 8.2 
& \underline{67.8} & 59.7 & 49.4 & 32.8 
& \textbf{\textit{40.4}} &  \textbf{\textit{6.6}} & \textbf{\textit{9.6}} &  \textbf{\textit{2.26}}
\\
& Troika 
& 48.8 & 18.7 & 20.1 & 7.2
& 66.4 & \underline{61.2} & 47.8 & 33.0
& \textit{\underline{37.4}} &  \textit{4.5}&  \textit{6.0} &  \textit{1.11}
\\
\rowcolor{gray!15}
\cellcolor{white}{}
& DCDA[PRG]
& \underline{54.6} & \underline{27.3} & \underline{25.8} & \underline{11.5}
& \textbf{68.6} & 56.4 & \underline{51.2} & \underline{33.8}
& \textit{35.5} &  \textit{4.4} &  \textit{6.7} &  \textit{1.30}
\\
\rowcolor{gray!15}
\cellcolor{white}{}
& DCDA[PRG+N]
& \textbf{55.0} & \textbf{27.7} & \textbf{26.7} & \textbf{12.0}
& \underline{67.8} & \textbf{62.5} & \textbf{51.4} & \textbf{35.8}
& \textit{35.3} &  \textit{\underline{6.4}}&  \textit{\underline{8.5}} &  \textit{\underline{1.76}}
\\
\hline
\end{tabular}
}
\caption{Overall Results ($\%$) on three benchmarks. In each setting, the best results are in bold and the second best are underlined. We report DFSP in its $t2i$ setting. Numbers in italics mean the results implemented with CLIP-base, others are with CLIP-large.
}
\label{tab:overall_results}
\end{table*}

\section{Experiments}


\subsection{Experimental Settings}
\noindent\textbf{Datasets and Metrics.}
We experiment with three popular benchmarks for CZSL: MIT-States \cite{isola2015discovering}, UT-Zappos \cite{yu2014fine} and C-GQA \cite{naeem2021learning}, 
and follow \cite{purushwalkam2019task,naeem2021learning} to split the data for training, validation and testing.
For each dataset, we compute the prediction accuracy of seen and unseen compositions, and report four metrics: the best Seen (\textbf{S}) and Unseen (\textbf{U}) accuracy, the best harmonic mean (\textbf{H}), and the Area Under the accuracy Curve (\textbf{AUC}). Among them, AUC is the most comprehensive one and is widely adopted as the core metric by previous works \cite{purushwalkam2019task,caila2024}.
Please see Appendix~\ref{sec:appendix_exp} for more datasets and metrics details.

\noindent\textbf{Closed World and Open World Settings.}
Given the attribute set $\mathcal{A}$ and object set $\mathcal{O}$, the complete compositional label set $\mathcal{C}$ should be the Cartesian product of $\mathcal{A}$ and $\mathcal{O}$, i.e., $\mathcal{C} = \mathcal{A} \times \mathcal{O}$ with size $|\mathcal{A}| \times |\mathcal{O}|$.
However, current benchmarks often operate in a \textit{closed world} setting where unseen compositions $\mathcal{C}_{u}$ in $\mathcal{C}_{te}$ are a small subset of $\mathcal{C} \setminus \mathcal{C}_s$ and are assumed to be known.
For example, MIT-States contains 28,175 possible compositions with 115 attributes and 245 objects, but the label space for testing is limited to 1,962 compositions (1,262 seen and 700 unseen), covering less than 7\% of the complete set.
Thus, we follow \cite{mancini2021open} to evaluate our model in the \textit{open world} setting, where the testing images remain unchanged but the testing label space is all possible combinations, i.e., $\mathcal{C}_{te} = \mathcal{C}$ and $\mathcal{C}_u = \mathcal{C} \setminus \mathcal{C}_s$, which is more challenging as the models have to generalize from a small set of seen to a very large set of unseen compositions.
Notably, not all the combinations are feasible, such as \textit{eroded cat}, for this, we apply post-training calibration \cite{csp2023,gipcol2024} to filter out unreasonable compositions.

\noindent\textbf{Baselines and Model Variants.}
We mainly compare our DCDA with the existing CLIP-based CZSL methods, including the vanilla CLIP without fine-tuning, CSP \cite{csp2023}, HPL \cite{hpl2023}, DFSP \cite{dfsp2023}, CAILA \cite{caila2024}, and Troika \cite{huang2024troika}.
We also include two non-CLIP-based baselines that are most similar to us, namely CGE \cite{naeem2021learning} and ADE \cite{hao2023learning}.

In V-Adapters, we propose a novel primitive-relevance guided (PRG) sampling method to select representative auxiliary compositions to sample rather than performing purely random (RD) sampling over all neighboring compositions. For detailed comparisons, we develop a variant DCDA[RD] and denote the vanilla model as DCDA[PRG].
With the imbalanced sample distribution, we also focus on the tail compositions in the neighbor set, and sample them according to the reciprocal of their image numbers (N) to supply DCDA[PRG], leading to a new variant DCDA[PRG+N], these two sampling strategies are switched batch by batch.
Please see Appendix~\ref{sec:appendix_exp_imple} for more implementation details.

\subsection{Main Results}

\textbf{Closed-world Performance.} As shown in Table~\ref{tab:overall_results} (top), our DCDA variants achieve state-of-the-art performance across benchmarks. On MIT-States, both DCDA[PRG] and DCDA[PRG+N] surpass previous methods with \textbf{over 3\% improvements} in AUC and harmonic mean (H), and \textbf{over 6\% gains} in seen accuracy compared to CAILA. For UT-Zappos, DCDA[PRG+N] \gyx{is the best}
on three metrics and achieves \textbf{substantial AUC improvements} over CGE \gyx{despite the inferior result on H}.
\gyx{The} initial experiments showed suboptimal performance on C-GQA due to its low-\gyx{quality} images, unfreezing CLIP's image encoder eventually enabled DCDA to achieve competitive second-place results, \gyx{see Appendix~\ref{sec:appendix_exp_cgqa} for more.}

\gyx{\textbf{Model Variants Analyses.}  
The significant margin between DCDA[PRG] and DCDA[RD] on all datasets validates our primitive-relevance guided sampling strategy. Especially, the improvement on MIT-States is promising, since the number of compositions surrounding its each primitive is more imbalanced than that of UT-Zappos. 
Moreover, in contrast to the slight performance gap between DCDA[PRG+N] and DCDA[PRG] on MIT-States and C-GQA, DCDA[PRG+N] shows great superiority on UT-Zappos,
demonstrating superior handling of class imbalance (sample size std: UT-Zappos's 465 vs. MIT-States' 12 vs. C-GQA's 22).
 }

\textbf{Cross-method Insights.} Among CLIP-based methods, CAILA, Troika, and DCDA form the top tier by injecting adapters for visual disentanglement. Our approach outperforms both competitors on \gyx{both the general dataset MIT-States and the domain-specific dataset UT-Zappos with fewer trainable parameters through partial-layer adapter insertion (our last-3 layers vs. full layers in CAILA and Troika)}, demonstrating superior cross-domain 
adaptability, \gyx{and} preserving CLIP's generalization as evidenced by open-world results (Table~\ref{tab:overall_results}, bottom).
\gyx{The comparable results on C-GQA also motivate us to develop more advanced learning paradigms in the future.}

\begin{table*}[]
\centering
\resizebox{0.55\linewidth}{!}{
\begin{tabular}{l|cccc}
\hline
\multicolumn{1}{c|}{\multirow{1}{*}{\textbf{Models}}}
&  S &  U & H & AUC
\\
\hline
\textbf{Full Model} (e.g., DCDA[PRG])
& 57.3 & \textbf{55.1} & \textbf{43.2} & \textbf{26.9}
\\
\hline
\ \  w/o L-Adapters
& 55.9 & 54.7 & 42.2 & 26.1 
\\
\ \  w/o V-Adapters
& 44.9 & 46.9 & 33.8 & 17.1 
\\
\hline
\ \  L-Adapter w/o other compositions
& \textbf{57.5} & 54.6 & 43.0 & 26.7 
\\
\ \  V-Adapter w/o other compositions
& 44.5 & 46.2 & 33.7 & 17.0 
\\
\ \  L\&V-Adapter w/o other compositions
& 44.2 & 46.1 & 33.6 & 16.7 
\\
\hline
\end{tabular}
}
\caption{Adapter Analysis on MIT-States in \textit{closed world}.}
\label{tab:ab_adapters}
\end{table*}

\subsection{Effectiveness of Adapters}

\noindent\textbf{The whole Adapters.}
We evaluate the contribution of L-Adapter and V-Adapter by analyzing the performance drop when one of them is removed. Notably, when all L-Adapters (reps. V-Adapters) are removed, the text (resp. image) encoder will act like CLIP's default frozen encoders to output an entangled representation for each input prompt (resp. image). We conduct experiments on MIT-States dataset under the \textit{closed world} setting, the results are shown in the second and third lines of Table~\ref{tab:ab_adapters}. We can see that the performance both declines when L-Adapters or V-Adapters are removed, indicating that they two both have a positive contribution to the DCDA model and are complementary to each other.
We also observe that the performance decrease of removing V-Adapters is greater than that of removing L-Adapters, which is consistent with our statement: textual primitive features are less entangled than visual ones, and the independent textual primitive features can still be captured by setting individual primitive prompts. 

\begin{figure*}[htbp]
\centering
\subfigure{
\begin{minipage}[c]{0.5\linewidth}
\centering
\includegraphics[width=6.5cm]{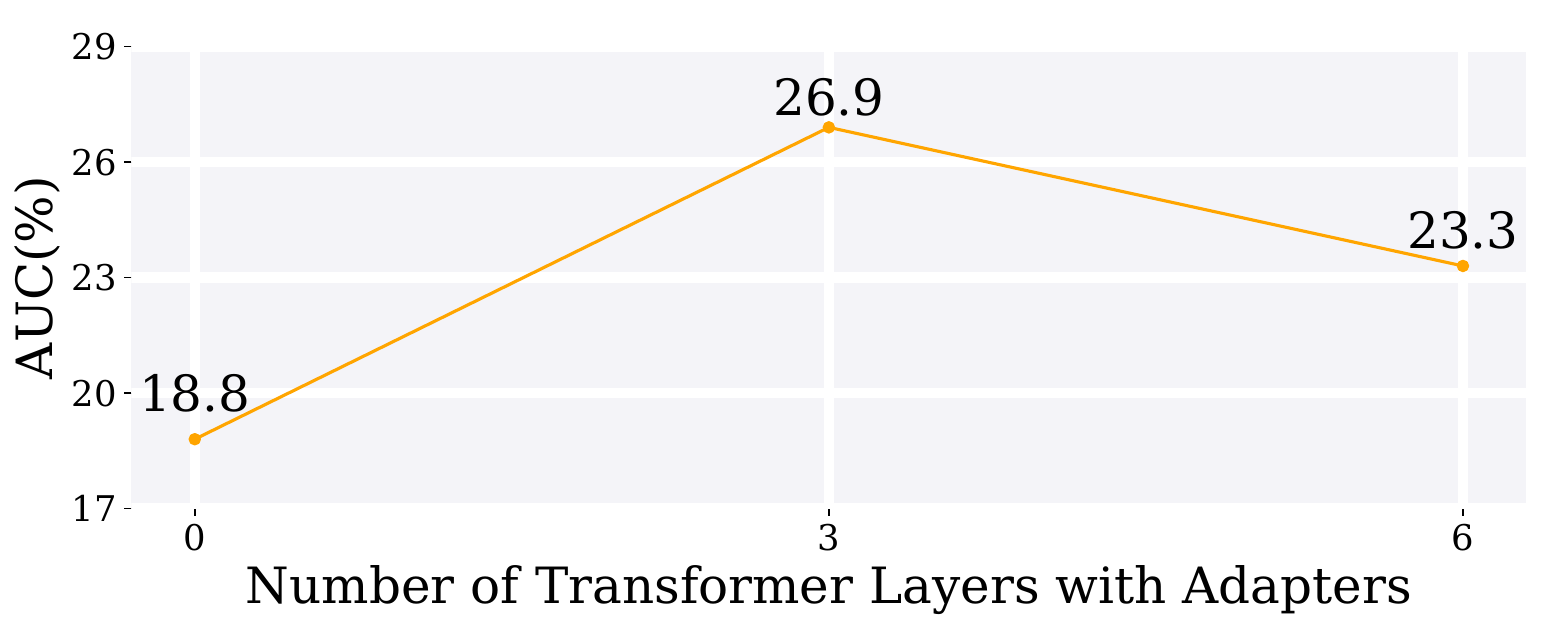}
\end{minipage}%
}%
\subfigure{
\begin{minipage}[c]{0.5\linewidth}
\centering
\includegraphics[width=6.5cm]{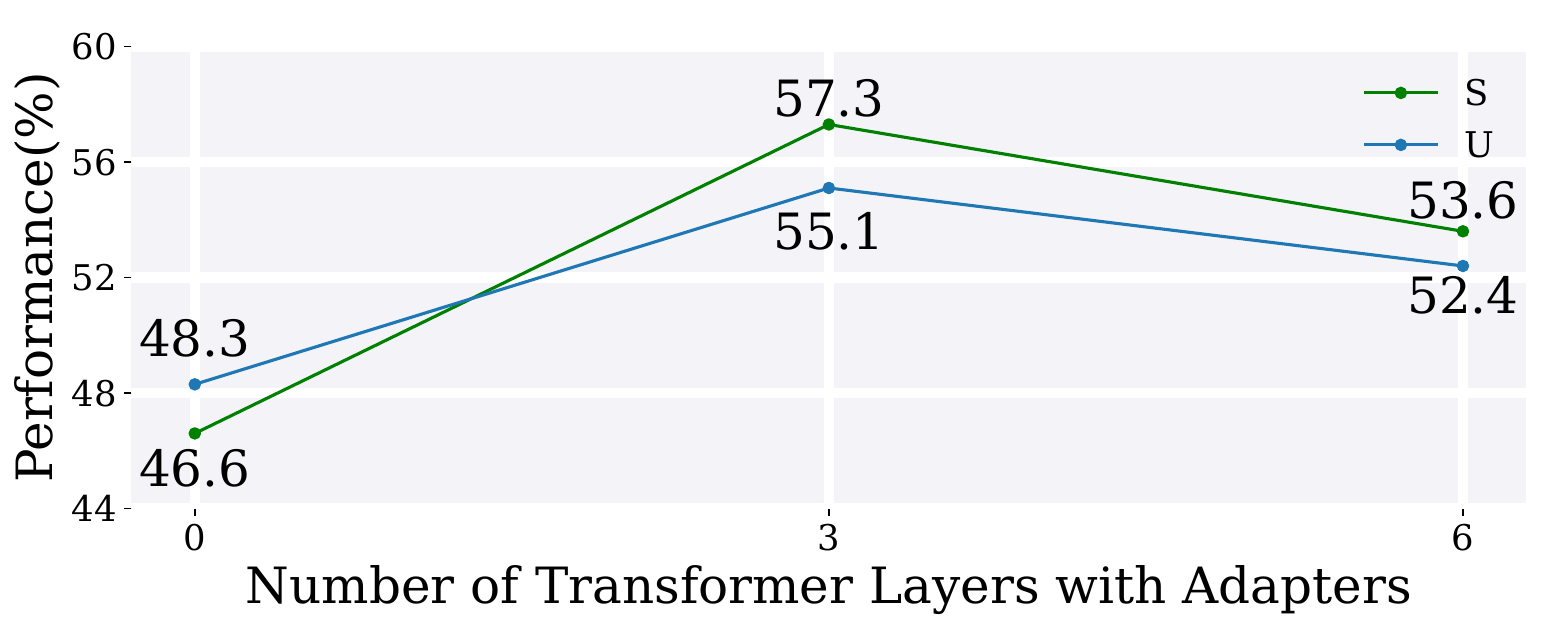}
\end{minipage}%
}%
\caption{Performance (AUC, S and U) of increasing the number of transformer layers with adapters on MIT-States.
}
\label{fig:ab_depth}
\end{figure*}

\noindent\textbf{Cross-composition Information in Adapters.}
We further validate the effectiveness of introducing primitive-sharing compositions for each target composition by deleting the compositional graph in L-Adapters and/or auxiliary compositions in V-Adapters.
Concretely, we merely keep the prompts' token embeddings trainable in L-Adapters; and/or replace the two auxiliary images with the target image itself in V-Adapters, which thus turns into the self-attention on the input image, but still projects the primitive features into different subspaces with different attention networks. The results on MIT-States are shown in the last three lines of Table~\ref{tab:ab_adapters}.
The performance drop indicates the superiority of introducing primitive-sharing compositions in both text and image encoders. In particular, we find that V-Adapters without primitive-sharing compositions even perform worse than removing the whole V-Adapters, illustrating that these compositions play a considerable role in constraining the learning of primitive features in different subspaces. In contrast, L-Adapters without neighboring (primitive-sharing) compositions perform better than removing the whole L-Adapters. However, the performance gap is slight, which may be attributed to that we also set the embeddings of tokens in the prompts tunable, the cross-composition primitive features can be implicitly captured by optimizing the token embeddings with multiple samples.

\subsection{Ablation Studies}\label{sec:ab}

\noindent\textbf{Insertion Location of Adapters.}
There are two choices for inserting adapters into a transformer block, i.e., inside or outside.
Therefore, we experiment with four configurations where every L-Adapter and V-Adapter are added inside or outside a transformer block, as shown in Table~\ref{tab:ab_location}.
Notably, the inside insertion includes adding after the self-attention and feed-forward layers in a block, i.e., the positions \ding{174} and \ding{173} in Figure~\ref{fig:framework}(c).
From Table~\ref{tab:ab_location}, it can be seen that adding V-Adapters outside transformer layers often achieves better performance no matter where L-Adapters are located, while there is no significant difference when shifting L-Adapters at different positions. This may be because when adding a V-Adapter inside the transform layer, there is no nonlinear transformation like Feed-Forward layer between two attention operations for extracting more informative features.
Regarding the better performance with inside L-Adapters and outside V-Adapters, we finally apply this configuration to three benchmarks.

\begin{table}[]
\centering
\resizebox{0.95\linewidth}{!}{
\begin{tabular}{cc|cc|cccc}
\hline
\multicolumn{2}{c|}{\textbf{L-Adapters}} & \multicolumn{2}{c|}{\textbf{V-Adapters}} 
& \multicolumn{4}{c}{\textbf{MIT-States}}
\\
I & O & I & O &  S &  U & H & AUC 
\\
\hline
\checkmark & &  \checkmark &
& 53.5 & 53.4 & 40.6 & 24.0 
\\
\checkmark & & & \checkmark
& \textbf{57.3} & 55.1 & \textbf{43.2} & \textbf{26.9}
\\
& \checkmark & \checkmark &
& 53.1 & 53.0 & 41.0 & 24.0 
\\
& \checkmark & & \checkmark
& 56.0 & \textbf{55.5} & \textbf{43.2} & 26.7

\\
\hline
\end{tabular}
}
\caption{Ablation Study on Adapters' Insertion Locations -- inside (I) or outside (O) one transformer block.}
\label{tab:ab_location}
\end{table}

\noindent\textbf{Insertion Depth of Adapters.}
We further show the performance change when we increase the number of transformer blocks with trainable adapters in Figure~\ref{fig:ab_depth}.
As mentioned earlier, we start from the top transformer layers.
It is clear that the best performance is achieved with the last 3 layers, while the last 6 layers may overfit the training data, with less generalization knowledge from our adapters.


\section{Conclusions and Outlook}
We present DCDA, a graph-guided framework that enhances CLIP for compositional zero-shot learning through dual adapters. Our key innovations include: 1) L-Adapters that aggregate textual primitives via compositional graph propagation, addressing label-side entanglement; 2) V-Adapters that extract invariant visual patterns through cross-attention and primitive-relevance guided sampling, effectively mitigating image feature entanglement.
Experiments across three benchmarks validate that DCDA learns more discriminative \gyx{and generalizable} primitive representations. This underscores the critical role of visual disentanglement in CZSL, as visual primitives exhibit stronger composition-dependent entanglement than textual counterparts. In the future, it is expected to apply our dual adapters in other vision-language tasks and develop dynamic graph construction mechanisms to handle open-vocabulary primitive discovery.

\section*{Limitations}

While our proposed DCDA demonstrates strong performance in compositional zero-shot learning (CZSL), several limitations warrant discussion as follows:
(1) the computational overhead of constructing and updating the compositional graph grows with the scale of attributes and objects, which may pose challenges for applications requiring real-time inference on large combinatorial spaces.
(2) while our primitive-relevance guided sampling mitigates data imbalance, extreme long-tailed distributions of attributes or objects (e.g., rare primitives with few compositions) may still lead to suboptimal disentanglement due to insufficient cross-composition supervision. Addressing these limitations could further enhance the robustness and scalability of our framework.



\section*{Acknowledgments}
This work is partly funded by the Zhejiang Provincial Natural Science Foundation of China under Grant No. LQ24F020034, and the Primary R\&D Plan of Zhejiang under Grant No. 2023C03198.
Jiaoyan Chen is mainly supported by the EPSRC project OntoEm (EP/Y017706/1). Xiang Chen is mainly supported by the Scientific Research Starting Foundation of Nanjing University of Aeronautics and Astronautics
(No.1015-YAH24096), and the High Performance Computing Platform of Nanjing University of Aeronautics and Astronautics.


\bibliography{custom}

\appendix

\section{Supplementary Methodology Details}
\begin{table*}[htbp]
\centering
\resizebox{0.6\linewidth}{!}{
\begin{tabular}{c|c|cc|cc|cc}
\toprule[0.5pt]
\multicolumn{1}{c|}{\multirow{2}{*}{\textbf{Datasets}}}
& 
\multicolumn{1}{c|}{\textbf{Composition}}
& 
\multicolumn{2}{c|}{\textbf{Train}} & \multicolumn{2}{c|}{\textbf{Validation}} 
& \multicolumn{2}{c}{\textbf{Test}}
\\
& $|\mathcal{A}|$ / $|\mathcal{O}|$ & $|\mathcal{C}_{s}|$
& $|\mathcal{X}_{s}|$
& $|\mathcal{C}_{s}|$ / $|\mathcal{C}_{u}|$ &  $|\mathcal{X}_{val}|$ &  $|\mathcal{C}_{s}|$ / $|\mathcal{C}_{u}|$ & $|\mathcal{X}_{te}|$
\\
\midrule[0.3pt]
MIT-States & 115 / 245 
& 1,262 & 30,338
& 300 / 300 & 10,420
& 400 / 400 & 12,995
\\
UT-Zappos & 16 / 12
& 83 & 22,998
& 15 / 15 & 3,214 
& 18 / 18 & 2,914 
\\
C-GQA & 413 / 674
& 5,592 & 26,920
& 1,040 / 1,252 & 7,280
& 888 / 923 & 5,098
\\
\bottomrule[0.5pt]
\end{tabular}
}
\caption{Statistics of Datasets for CZSL. $|\mathcal{C}_u|$ here is the number of unseen compositions in the \textit{closed world} setting. 
}
\label{tab:dataset_statistics}
\end{table*}

\begin{table*}[]
\centering
\resizebox{0.5\linewidth}{!}{
\begin{tabular}{l|cccc}
\hline
\multicolumn{1}{c|}{\multirow{1}{*}{\textbf{Models}}}
&  S &  U & H & AUC
\\
\hline
Frozen Encoder + 3 V-Adapters (default)
&34.8  & 23.4 &21.6  & 6.9
\\
Extra Adapters + 3 V-Adapters
& 38.8 & 26.6 & 23.8 & 8.7
\\
Full Fine-Tuning  + 3 V-Adapters
& 38.5 & \textbf{28.8} & \textbf{25.3} & \textbf{9.4}
\\\hline
Full Fine-Tuning  + 1 V-Adapters
& 38.9 & 26.4 & 24.6 & 8.8
\\
Full Fine-Tuning  + 2 V-Adapters
& \textbf{39.2} & 27.3 & 24.9 & 9.0
\\
Full Fine-Tuning  + 4 V-Adapters
& 38.8 & 25.6 & 24.2 & 8.5
\\
Full Fine-Tuning  + 6 V-Adapters
&31.9  & 13.7 & 14.1 & 3.5
\\
\hline
\end{tabular}
}
\caption{Ablation on adding more trainable parameters in the vision encoder on C-GQA in \textit{closed world}. All results are tested with the ``[PRG+N]'' sampling method.}
\label{tab:ab_cgqa}
\end{table*}

\subsection{Computing the Sampling Probability}\label{sec:appendix_method_sampling_prob}
To sample the top-$n$ most and least relevant compositions according to the top-$n$ maximum and minimum primitive relevance scores, we take two ways to compute the sampling probabilities.
More specifically, the probability for selecting the top-$n$ most relevant compositions, e.g., $(a, o'_i)$, and the probability for selecting the top-$n$ least relevant compositions, e.g., $(a, o'_j)$, are computed as, respectively:
$$
\scalebox{0.85}{$
    P(o'=o'_i) = \frac{A^{\text{obj}}_{o,o'_i}}{\sum_{o'_k \in \{o'\}} A^{\text{obj}}_{o,o'_k}}, P(o'=o'_j) = -\frac{A^{\text{obj}}_{o,o'_j}}{\sum_{o'_k \in \{o'\}} A^{\text{obj}}_{o,o'_k}},$}
$$
where $A^{\text{obj}}_{o,o'_i}$ means the relevance score between objects $o'_i$ and $o$, the same applies to the others.
The difference is that the normalized minimum scores are multiplied by $-1$ to ensure that the top-1 least relevant composition has the highest probability of being sampled among the $n$ least relevant compositions.
Consider the example in Figure~\ref{fig:framework}(b), \textit{red wine, red rose} and \textit{red leaf} are top-3 least relevant compositions for \textit{red tomato}, with top-3 normalized minimum object-relevance scores: $0.1, 0.2$ and $0.2$, respectively, while \textit{red wine} is the top-1 least relevant composition with the highest sampling probability among these three compositions.

To avoid being confused by the positive and negative sampling probabilities, we divide the representative compositions into two groups and sample over the top-$n$ most and least relevant compositions independently.
To be more specific, we switch them batch by batch.

\section{Supplementary Experiment Details}\label{sec:appendix_exp}

\subsection{Dataset}
\underline{MIT-States} contains 53,753 real-world images, annotated by a variety of classes with 245 objects and their 115 attributes in the general domain. In \textit{closed world}, it provides 1,962 compositions in total, 1,262 of which are seen used for training, and 700 are unseen with 300 for validation and 400 for testing.
\underline{UT-Zappos} is a more domain-specific dataset, containing 50,025 images of shoes paired with their material attributes. In total, it has 16 attributes and 12 objects, yielding 83 seen and 33 unseen compositions under the \textit{closed world} setting. 
\underline{C-GQA}, derived from of Stanford GQA dataset \cite{hudson2019gqa}, is the most extensive dataset for CZSL, containing 7,767 compositions (5,592/2,175 as seen/unseen), 413 attribute classes, 674 object classes, and 39,298 images in total.
Table~\ref{tab:dataset_statistics} summarizes the dataset statistics in the \textit{closed world} setting, the \textit{open world} has the same set of testing images, i.e., $\mathcal{X}_{te}$, but larger candidate label set, i.e., all possible compositions $\mathcal{C}$ obtained by the Cartesian product of $\mathcal{A}$ and $\mathcal{O}$.

\subsection{Evaluation Metrics}
We compute the prediction accuracy for recognizing seen and unseen compositions, i.e., the generalized CZSL, in both \textit{closed world} and \textit{open world} scenarios.
Specifically, due to the inherent bias towards seen classes, we follow the current standard \cite{purushwalkam2019task,csp2023} to add a scalar bias to the prediction scores of unseen classes and vary the bias from $-\infty$ to $+\infty$ to get a seen-unseen accuracy curve, which indicates the seen accuracy on the x-axis and unseen accuracy on the y-axis.
As a result, we report the best seen accuracy (\textbf{S}), where the bias is set to $-\infty$ and the models only predict on the seen labels, and report the best unseen accuracy (\textbf{U}), where the bias is set to $+\infty$ and the models only predict on the unseen labels.
We also calculate the best harmonic mean (\textbf{H}), where a harmonic mean value is first computed for each point on the curve to balance the seen accuracy ($acc_s$) and unseen accuracy ($acc_u$) as $(2 \times acc_s \times acc_u)/(acc_s+acc_u)$, and then the highest value across all the selected points is reported.
Finally, we compute the Area Under the accuracy Curve (\textbf{AUC}) as a comprehensive metric.

\subsection{Implementation Details}\label{sec:appendix_exp_imple}
We implement our models with PyTorch and use Adam as the optimizer, with the learning rate set to 5e-5, 5e-5, 1e-5, and the batch size set to 32, 32, 16 for MIT-States, UT-Zappos, C-GQA, respectively, the weight decay is set to 5e-5 for all datasets.
The GNN module is implemented as GCN with $K=2$.
$n$ is set to 5 for selecting the representative auxiliary compositions.
The initialized values of $\alpha, \beta, \gamma$ are all set to 1, and then optimized together with other parameters.
The optimal hyperparameter configurations are determined using AUC on the validation set.
All the experiments are run on a single NVIDIA Tesla A100 GPU with 40GB memory.

\subsection{Experiments on C-GQA}\label{sec:appendix_exp_cgqa}

To deal with the low-resolution and small-size images in C-GQA, we tried to add more trainable parameters, such as inserting more adapters in CLIP's image encoder or fully fine-tuning the whole image encoder, to adapt CLIP to this kind of image.
More specifically, we add a downsample-upsample style adapter, which is similar to the adapter used in CAILA and Troika, after the self-attention layer and feed-forward layer in each vision transformer block (i.e., the positions \ding{174} and \ding{173} in Figure~\ref{fig:framework}(c)), except for the last three transformer blocks where our V-Adapters have already been there.
In addition, we also try to unfreeze the whole image encoder to fully update its parameters, where our V-Adapters are added in the last three layers for feature disentanglement.
The results are presented in the second and third lines of Table~\ref{tab:ab_cgqa}, respectively.
Moreover, since a single A100 GPU with 40GB memory cannot afford these extra adapters or fully fine-tuned parameters with CLIP-large that contains 24 vision transformer (ViT) layers, we instead use CLIP-base with ViT-B/32 as its image encoder, and re-run baselines for a fair comparison.
Regarding that some baselines with CLIP-large still perform worse than our methods with CLIP-base, we omit re-implementing them to save computation costs.

From Table~\ref{tab:ab_cgqa}, we can see that introducing more trainable parameters indeed achieves superior performance, in comparison with the vanilla models that only include 3 V-Adapters in the last three layers of the frozen image encoder.
Especially, the fine-tuning method performs better.
As a result, we implement our DCDA[PRG] and DCDA[PRG+N] with the whole image encoder fully tunable, the resulting models together with our V-Adapters achieve very competitive performance on C-GQA compared with the SOTA CAILA.
Moreover, we also vary the number of transformer layers with V-Adapters inserted, starting from the top transformer layers, the results are as shown in the last four lines of Table~\ref{tab:ab_cgqa}. From Table~\ref{tab:ab_cgqa}, we have similar observations as in Figure~\ref{fig:ab_depth}, i.e., too few (e.g., only one) V-Adapter is not enough to disentangle the primitive features, while too many (e.g., 6) V-Adapters may overfit to disentangle the training data, resulting in poor generalization.
To sum up, the fine-tuned image encoder and three V-adapters lead to a balance between adapting CLIP to the C-GQA dataset and disentangling its image features.

\section{Supplementary Case Studies}\label{sec:appendix_case_study}

\begin{figure}[htbp]
\centering
\subfigure{
\begin{minipage}[c]{0.8\linewidth}
\centering
\includegraphics[width=4.5cm]{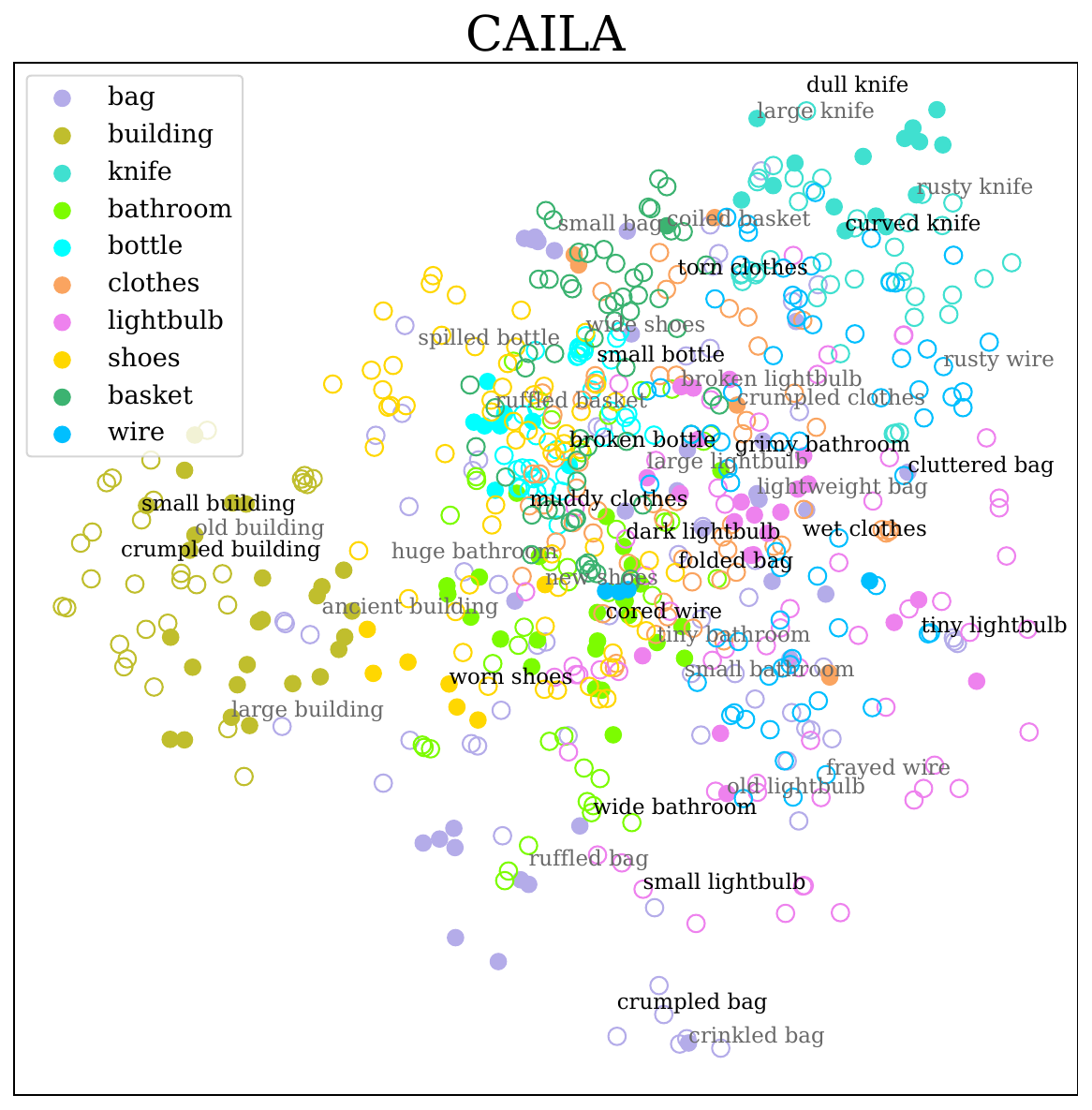}
\end{minipage}%
}%
\\
\subfigure{
\begin{minipage}[c]{0.8\linewidth}
\centering
\includegraphics[width=4.5cm]{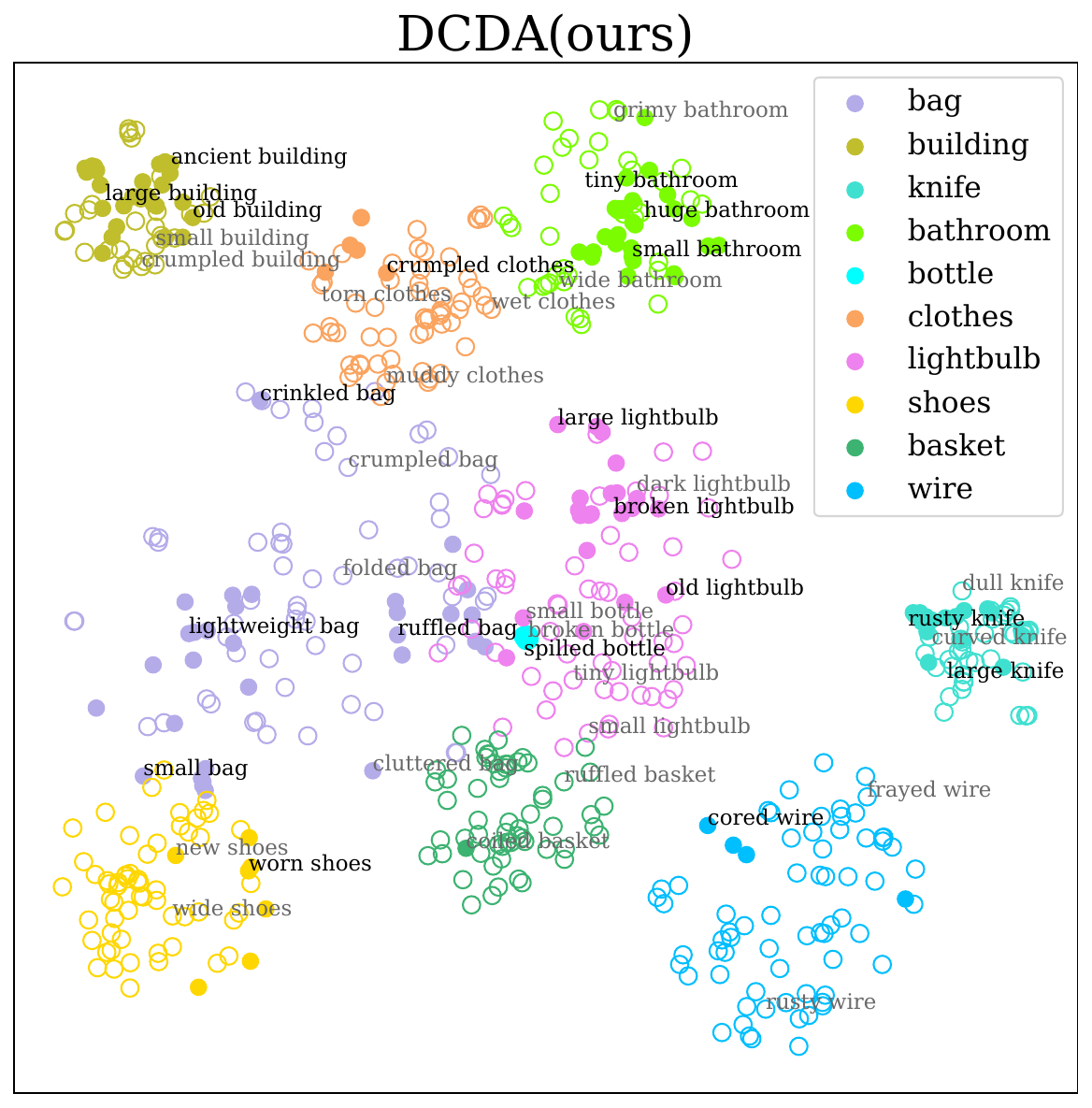}
\end{minipage}%
}%
\centering
\caption{$t$-SNE visualizations of disentangled object representations of images in the test set of MIT-States, learned by CAILA \cite{caila2024} and our DCDA. Solid and hollow circles represent images of seen and unseen compositions, respectively. Best viewed in color.
}\label{fig:disentangled_object_visualization}
\end{figure}

We use examples from MIT-States to analyze the disentanglement of primitive features learned by CAILA and our DCDA (e.g., DCDA[PRG]), especially those visual ones.
Specifically, we first randomly sample a set of seen and unseen compositions from the test set of MIT-States whose annotated attributes or objects have high diversity.
Here, we use the number of associated objects (resp. attributes) to roughly measure the diversity of an attribute (resp. object) as a wider range of objects (resp. attributes) would lead to more diverse appearances of attributes (resp. objects). For example, in Figure~\ref{fig:disentangled_attribute_visualization}, attribute \textit{broken} describes 40 objects in the training set ranging from \textit{car}, \textit{drum} to \textit{furniture} with different damaged states; similar to the object \textit{knife} in Figure~\ref{fig:disentangled_object_visualization}, which is paired with 9 attributes in the training set.
In addition, we also manually select $2\sim3$ attributes (resp. objects) whose associated objects (resp. attributes) are fewer but look greatly different, e.g., the attribute \textit{worn} in Figure~\ref{fig:disentangled_attribute_visualization}.

Then, for each sampled composition, we randomly extract at least 3 testing images and visualize their visually disentangled attribute and object representations learned by our DCDA and CAILA in Figure~\ref{fig:disentangled_attribute_visualization} and Figure~\ref{fig:disentangled_object_visualization}, respectively, where different colors indicate different attribute or object labels.
More specifically, for DCDA, we extract the features learned by our V-Adapters, i.e., $\bm{H}_{\ast \rightarrow A}^v$ and $\bm{H}_{\ast \rightarrow O}^v$; while for CAILA, we save the features learned by its attribute and object-specific vision encoding blocks.

From Figure~\ref{fig:disentangled_attribute_visualization} and Figure~\ref{fig:disentangled_object_visualization}, it can be seen that the attribute embeddings or object embeddings learned by our model are clustered into different groups w.r.t different attributes or objects in each vector space.
For example, in Figure~\ref{fig:disentangled_attribute_visualization}, \textit{broken car}, \textit{broken drum} and \textit{broken furniture} are in the same cluster in the attribute space, i.e., their learned attribute features are similar even though they show different \textit{broken} state w.r.t different objects, similar to \textit{curved knife} and \textit{large knife} in Figure \ref{fig:disentangled_object_visualization}.
Also, \textit{broken car} is a seen composition, while \textit{broken drum} and \textit{broken furniture} are two unseen compositions.
However, the attribute embeddings as well as object embeddings learned by CAILA scatter in the attribute and object space, respectively. \ul{All of these illustrate that our DCDA captured similar visual features specific to each primitive, which is discriminative and generalizable.}

Moreover, we also find that the attribute and object embeddings of the same composition are divided into different clusters with different neighbors in two spaces, for example, \textit{large knife} and \textit{dull knife} are two neighbors from the cluster of \textit{knife} in the object space, while they fall into the clusters of \textit{large} and \textit{dull} in the attribute space with neighbors \textit{large building} and \textit{dull brass}, respectively.
\ul{This indicates that our method indeed disentangles the attribute and object features into different representation spaces.}

\end{document}